\documentclass[acmtog]{acmart}
\usepackage[ruled,vlined]{algorithm2e}
\usepackage{hyperref}
\usepackage{bm}

\usepackage{bbding}
\RestyleAlgo{ruled}
\RestyleAlgo{vlined}

\usepackage{booktabs}

\SetKwComment{Comment}{// }{}

\usepackage{graphicx}
\usepackage{subcaption}
\usepackage{multirow}
\usepackage{makecell, multicol, diagbox}

\makeatletter
\newcommand\paragraphNew{\@startsection{paragraph}{4}{\parindent}%
  {-.5\baselineskip \@plus -2\p@ \@minus -.2\p@}%
  {-3.5\p@}%
  {\ACM@NRadjust{\@parfont}}}
	\makeatother
	
\AtBeginDocument{%
  \providecommand\BibTeX{{%
    \normalfont B\kern-0.5em{\scshape i\kern-0.25em b}\kern-0.8em\TeX}}}

\usepackage{colortbl}
\newcommand{\newrevised}[1]{#1}
\newcommand{\revised}[1]{#1}

\renewcommand\footnotetextcopyrightpermission[1]{}

\definecolor{gold}{RGB}{221, 196, 65}
\newcommand{\red}[1]{\cellcolor{red!40}{#1}}
\newcommand{\gold}[1]{\cellcolor{gold!40}{#1}}

\newcommand{\textred}[1]{\colorbox{red!40}{#1}}
\newcommand{\textgold}[1]{\colorbox{gold!40}{#1}}

\begin{document}

\makeatletter
\setcopyright{none}
\settopmatter{printacmref=false}

\renewcommand\footnotetextcopyrightpermission[1]{}
\def\@copyrightspace{}

\fancyfoot{}
\makeatother



\title{\textbf{SRUG}: \textbf{S}hadow-Guided \textbf{R}elightable \textbf{U}rban Scene with \textbf{G}eneration Model}

\author{Yonghao Zhao}
\affiliation{
 \institution{College of Computer Science, Nankai University}
 \country{China}
}
\email{applezyh@outlook.com}

\author{Zexin Yin}
\affiliation{
 \institution{College of Computer Science, Nankai University}
 \country{China}
}
\email{Zexin.yin.cn@gmail.com}

\author{Jian Yang}
\affiliation{
   \institution{Nankai University and Nanjing University}
   \country{China}
}
\email{csjyang@njust.edu.cn}

\author{Beibei Wang}
\affiliation{
   \institution{School of Intelligence Science and Technology, Nanjing University}
   \country{China}
}
\email{beibei.wang@nju.edu.cn}

\author{Jin Xie$^\dagger$}
\thanks{$^\dagger$Corresponding author.}
\affiliation{
   \institution{School of Intelligence Science and Technology, Nanjing University}
   \country{China}
}
\email{csjxie@nju.edu.cn}



\begin{abstract}
Creating relightable urban scenes from images or videos is widely useful but highly ill-posed. Urban environments are typically unbounded and extend beyond the visible regions. As a result, many portions of the scene remain unobserved, yet these invisible regions can cast shadows onto visible areas. Reasonably modeling shadows cast by such invisible regions is challenging and poses a significant obstacle to creating relightable urban scenes. At the same time, sparse input views and complex illumination conditions further complicate relighting, as they introduce severe ambiguities in material decomposition. In this paper, we propose \textbf{S}hadow-guided \textbf{R}elightable \textbf{U}rban Scene with \textbf{G}eneration model (SRUG), a novel framework designed to address relighting challenges in urban scenes. SRUG leverages shadows to guide a 3D completion model for recovering the geometry of invisible regions, promoting the synthesis of physically reasonable shadows. In addition, SRUG employs an iterative material decomposition scheme that applies the large material model (LMM) to provide material supervision and iteratively decompose the scene's material properties, enabling robust material decomposition. Building upon these components, we introduce a physically-based lighting model that captures the complex illumination of urban scenes and supports reliable relighting. Extensive quantitative evaluations and visual comparisons demonstrate that our method outperforms existing approaches in both novel view synthesis and relighting tasks.
\end{abstract}


\begin{teaserfigure}
\centering
\includegraphics[width=\textwidth]{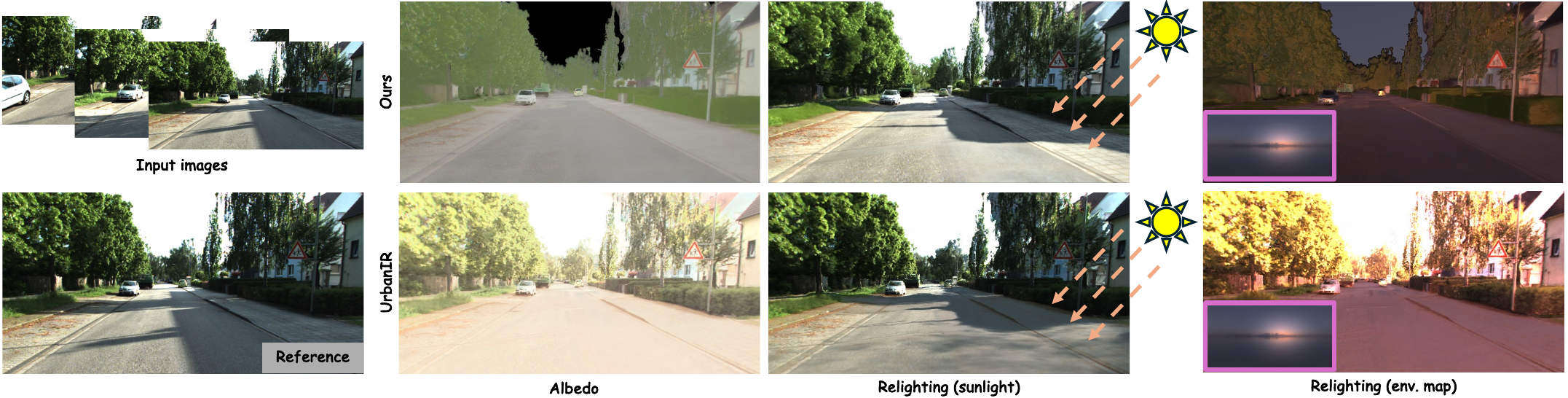}
\caption{
    We propose SRUG, a novel framework for constructing relightable urban scenes from multi-view images or videos. SRUG reconstructs 3D scene representations, enables robust material–lighting decomposition, and supports physically reasonable relighting. Compared with the existing urban scene relighting method UrbanIR~\cite{lin2023urbanir}, SRUG more effectively disentangles lighting from material properties, avoiding lighting bake-in in the estimated albedo. During relighting, SRUG produces physically consistent shadows under changes in sun direction and avoids floating artifacts. Moreover, under environment map–based relighting, SRUG yields more reliable results due to its robust scene decomposition.
}
\label{fig:teaser}
\end{teaserfigure}

\maketitle

\section{Introduction}
Reconstructing relightable 3D urban scenes from multiple images or videos has numerous applications, including building world models, aiding autonomous driving, and creating digital twins. However, this task is inherently difficult due to the ill-posed nature of separating lighting and materials from images or videos. The challenges are further intensified by the characteristics of urban environments. First, shadow modeling is essential for urban scene relighting. Yet, the unbounded nature of urban scenes leaves many regions unobserved. These invisible areas can still cast shadows onto visible regions, making shadow modeling particularly difficult. Moreover, urban environments exhibit complex illumination composed of sunlight, skylight, and indirect light, which complicates material–lighting decomposition and often leads to severe lighting bake-in artifacts. These challenges are further amplified under sparse-view settings.

Several studies~\cite{rudnev2022nerf-osr, wang2023FEGE, lin2023urbanir} have been developed to create relightable urban scenes by incorporating specially designed regularization priors into the neural radiance field (NeRF)~\cite{mildenhall2021nerf} framework. However, these methods face challenges related to their low expressive power and high computational demands, which significantly limit both the effectiveness and efficiency of relighting. Furthermore, while these regularization priors assist in the separation of materials and lighting, they are restricted to diffuse colors, which leads to noticeable lighting bake-in issues. Recent advancements in 3D Gaussian Splatting (3DGS)~\cite{kerbl20233d} show enhanced representation capabilities. Although several approaches~\cite{liang2024gsir, gao2025relightable, jiang2024gsshader, gu2024IRGS, du2024gsid} address object- or small-scene relighting, they neglect invisible regions and struggle with material decomposition under complex urban illumination, limiting their effectiveness in urban scenes. \newrevised{GS-ID~\cite{du2024gsid} leverages a large material model (LMM) for material decomposition and conventional shadow mapping for shadow modeling. However, it relies solely on the LMM, which limits the robustness of the decomposition, and it does not consider shadow modeling of invisible areas.}

In this paper, we propose \textbf{S}hadow-guided \textbf{R}elightable \textbf{U}rban Scene with \textbf{G}eneration model (SRUG), a novel framework for urban scene relighting. The core challenge of urban scene relighting lies in completing invisible regions, ensuring physically plausible shadow effects, and robustly decomposing scene materials under complex illumination and sparse-view conditions. To address the challenges posed by invisible regions, we propose a shadow-guided invisible geometry completion module. This module leverages shadows to guide a 3D completion model in recovering the geometry of invisible areas, thereby enabling physically plausible shadow effects. To facilitate shadow-based guidance, we introduce a differentiable Gaussian shadow mapping (DGSM), which models shadows in a differentiable manner and supports effective shadow-guided optimization. To achieve robust scene decomposition, we further employ a pretrained large material model to provide material supervision for scene material decomposition. However, directly applying LMMs to real-world scenes often produces unrealistic results due to the domain gap between the synthetic data used for LMM training and real-world urban application scenarios. \newrevised{To address this limitation, we propose an iterative material decomposition scheme that leverages the scene’s materials as additional conditioning to bridge the LMM domain gap, progressively refines material properties, and achieves robust decomposition.} We then apply a physically based lighting model for urban scenes that explicitly accounts for sunlight, skylight, and indirect illumination, enabling reliable relighting.

We evaluate our method on the real-world and synthetic datasets across novel view synthesis (NVS), material decomposition, and relighting tasks. Our framework consistently achieves state-of-the-art performance compared with existing relighting methods. To summarize, our main contributions are as follows:
\begin{itemize}
    \item a novel framework, SRUG, for constructing relightable urban scenes from multi-view images or videos,
    \item a shadow-guided invisible geometry completion module that leverages shadow cues to guide a 3D completion model in recovering invisible regions and synthesizing physically plausible shadows, and
    \item \newrevised{an iterative material decomposition scheme that uses LMM to refine material properties for robust decomposition.}

\end{itemize}

\section{Related Work}
\paragraphNew{Novel view synthesis} generates images from arbitrary viewpoints using image sequences or videos. Neural radiance fields~\cite{mildenhall2021nerf} pioneered scene representation through volume rendering with neural networks, achieving photorealistic synthesis. Subsequent works~\cite{muller2022instant, barron2022mipnerf360, tancik2022blocknerf, wang2021neus, pumarola2021d-d_nerf, poole2022dreamfusion, jin2023tensoir} have expanded NeRF's applications across diverse domains. To enhance efficiency and quality, 3D Gaussian splatting~\cite{kerbl20233d} introduced a radiance field representation using 3D Gaussians, achieving state-of-the-art results through efficient rasterization. This advancement has inspired numerous extensions, including improvements to synthetic quality~\cite{lu2024scaffold, yu2024mip-splatting, cheng2024gaussianpro}, geometry reconstruction~\cite{huang20242dgs, yu2024gaussian-gof, zhang2024rade}, autonomous driving scenes~\cite{yan2024street, zhou2024drivinggaussian}, dynamic scenes~\cite{Wu_2024_CVPR_4DGS, huang2024sc}, and inverse rendering~\cite{liang2024gsir, gao2025relightable, gu2024IRGS, du2024gsid, chen2025invrgb+}.

\paragraphNew{Object-scale inverse rendering} is a fundamental task in computer graphics and vision, aiming to estimate scene properties—geometry, materials, and lighting—from images or videos. Recent advances in neural rendering have enhanced inverse rendering by enabling joint estimation of these properties through differentiable rendering frameworks. Such methods improve the accuracy of property decomposition while maintaining high-quality rendering. Among these approaches, NeRF-based methods~\cite{jin2023tensoir, li2024tensosdf, liu2023nero, boss2021nerd, boss2021neural, yao2022neilf, zhang2023neilf++, srinivasan2021nerv} leverage Neural Radiance Fields to achieve inverse rendering, demonstrating the ability to recover scene properties while preserving view-consistent rendering. Other methods~\cite{liang2024gsir, gao2025relightable, jiang2024gsshader} integrate 3D Gaussian Splatting (3DGS) into inverse rendering, significantly improving rendering efficiency and quality.

\paragraphNew{Large-scale inverse rendering} Building on advances in object-scale inverse rendering, NeRF-OSR~\cite{rudnev2022nerf-osr, Haiyang2025GaRe} models urban scene properties by learning from in-the-wild data captured under varying lighting conditions. FEGR~\cite{wang2023FEGE} utilizes SDF to reconstruct the precise geometry of the scene and extracts meshes for shadow calculation. UrbanIR~\cite{lin2023urbanir} introduces material regularization to improve the quality of decomposition and utilizes a shadow prediction model as a prior to enable shadow-based optimization. \newrevised{GS-ID~\cite{du2024gsid} uses an LMM for material decomposition, but it does not fully address the model’s domain gap, as the LMM is primarily used for basic material supervision. It also employs conventional shadow mapping, yet does not account for shadows cast by invisible regions. Differentiable shadow mapping~\cite{worchel2023differentiable} makes shadow-based optimization possible. It extends improved shadow mapping techniques~\cite{donnelly2006variance_vsm, gumbau2011shadow_gsm} into differentiable formulations, but they are mainly designed for mesh representations. Inv-RGB+L~\cite{chen2025invrgb+} improves decomposition and relighting accuracy by incorporating LiDAR data; however, its reliance on specialized hardware limits its applicability compared with purely vision-based approaches.} In addition to overfitting approaches, several methods~\cite{philip2019multi, griffiths2022outcast, zeng2024rgb2x, DiffusionRenderer, kocsis2024intrinsic, luo2024intrinsicdiffusion, he2025unirelight} adopt end-to-end inverse rendering based on neural networks. Trained on extensive data, these methods learn to capture intrinsic material and lighting properties directly from images or videos. However, they typically do not provide the explicit 3D scene representation, which limits their generalizability.
\begin{figure*}
    \begin{center}
        \includegraphics[width=0.95\textwidth]{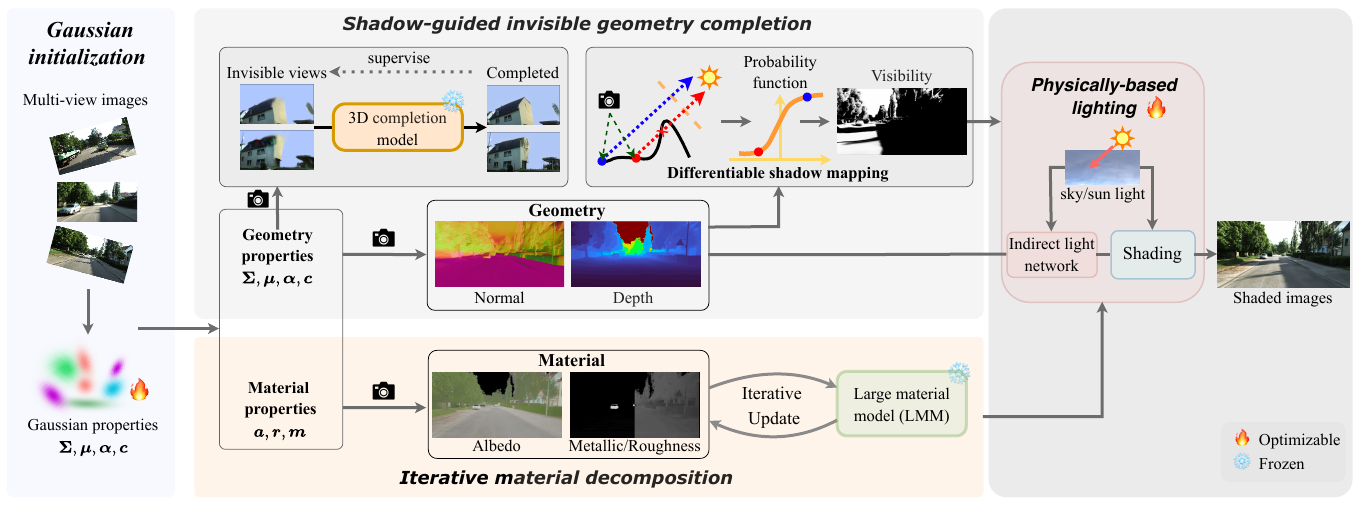}
    \end{center}
        \caption{\textbf{Overview of the SRUG framework.} We first initialize a Gaussian-based scene representation, including the geometry and appearance of visible regions within the scene. Based on the initialized Gaussians, we construct relightable urban scenes through two key components:  (1) a \textbf{shadow-guided invisible geometry completion module}. It employs differentiable shadow mapping to use shadows as supervisory signals for guiding a 3D completion model to recover invisible regions. This design enables physically plausible shadow synthesis; and  (2) an \textbf{iterative material decomposition scheme}, which utilizes a large material model to progressively refine scene material properties, enabling robust material decomposition. Building on these components, we introduce a physically-based lighting model for urban scenes that explicitly models sunlight, skylight, and indirect light to enable reliable relighting.}
        \label{fig:pipeline}
\end{figure*}

\section{Method}

\subsection{Overview of SRUG framework}
Our work aims to design a framework for reconstructing scene geometry and decomposing material properties from multi-view urban images or videos, ultimately enabling urban scene relighting via physically-based rendering. To this end, we introduce a shadow-guided invisible geometry completion module (see Sec.~\ref{sec:shadow}). By leveraging shadow-guided 3D completion models, our approach recovers the geometry of invisible areas and enables physically plausible shadow synthesis. Building on the completed geometry, we further decompose scene materials to support relighting. To achieve robust material decomposition, we employ a large material model to provide material supervision and iteratively refine the scene’s material properties (see Sec.~\ref{sec:material}). In addition, we introduce a physically-based lighting model that explicitly models sunlight, skylight, and indirect light, enabling reliable relighting under complex urban environments (see Sec.~\ref{sec:base}). Finally, we describe the overall optimization strategy of our framework (see Sec.~\ref{sec:training}).

\subsection{Shadow-guided invisible geometry completion}
\label{sec:shadow}
Urban scenes contain numerous occluding elements (e.g., buildings, vehicles) that create complex shadows under sunlight, making accurate shadow modeling essential for relighting. However, the unbounded nature of urban environments introduces significant challenges, particularly for capturing shadows cast by geometry outside the visible scene. To address this issue, we introduce a shadow-guided invisible geometry completion module. This module employs a pretrained 3D completion model to provide visual supervision for reconstructing invisible regions. However, relying solely on visual supervision lacks physical consistency. To ensure physically plausible shadows, we further propose a differentiable shadow mapping, which enables shadow-guided optimization, allowing shadows to serve as physically grounded supervision.

Specifically, we use Rade-GS~\cite{zhang2024rade} as the 3D representation baseline, which extends 3DGS with accurate depth and normal modeling. And we use a standard NVS training strategy to reconstruct the geometric structure of visible areas in the scene from multiple views. Based on this, we employ a pretrained 3D completion model, Difix3D~\cite{wu2025difix3d+}, to visually complete invisible areas. Difix3D repairs a target-view image conditioned on a given reference image. We exploit this capability to generate pseudo-supervision for invisible regions. Given a training view with camera parameters \(c\) and image \(I_{\text{gt}}\), we randomly perturb the camera pose (via translation and rotation) to obtain a novel camera \(c'\). Using \(c'\), we render an image \(I_{\text{gt}}'\), which typically covers regions that are invisible in the original training views. We then use the original image \(I\) as the reference and apply Difix3D to repair the rendered image \(I_{\text{gt}}'\). Leveraging the strong completion capability of Difix3D, the invisible areas in \(I_{\text{gt}}'\) are effectively repaired. The repaired image, together with the novel camera, is incorporated into the novel-view dataset and used as supervision for reconstructing invisible regions.

\begin{figure}[t!]
    \begin{center}
        \includegraphics[width=0.95\linewidth]{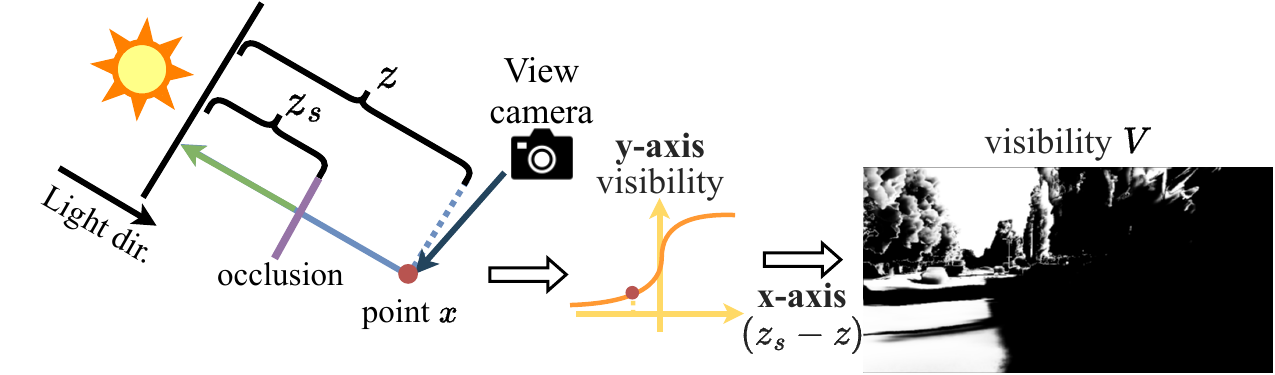}
    \end{center}
        \caption{
         \revised{In contrast to standard shadow mapping, DGSM replaces the binary comparison with a differentiable formulation. Given a shading point $x$, we compute its depth $z$ and corresponding shadow map depth $z_s$. Instead of the hard comparison $V = (z_s > z)$, we define visibility using a probability function, $V = \text{Sigmoid}\big(k(z_s - z)\big)$, which enables visibility for optimization.}
        }
        \label{fig:shadow}
\end{figure}

\revised{Difix3D provides pseudo-supervision for completing invisible regions; however, this completion relies purely on completion capability and lacks physical consistency. To address this limitation, we introduce shadows as physical guidance to provide more grounded supervision. Incorporating shadow-based supervision requires differentiable shadow modeling. Although Gaussian-based ray tracing can estimate visibility by accumulating opacity along rays, it increases computational overhead in large-scale outdoor scenes. In our setting, where sunlight can be approximated as a directional light, shadow mapping offers a more efficient alternative. However, the hard binary depth comparison in standard shadow mapping breaks differentiability and prevents gradient-based optimization. To this end, we propose the DGSM, as illustrated in Fig.~\ref{fig:shadow}. DGSM is inspired by improved shadow mapping techniques~\cite{donnelly2006variance_vsm, gumbau2011shadow_gsm}, which convert binary depth tests into continuous probabilities based on neighborhood depth statistics. In a similar spirit, DGSM reformulates hard depth comparisons into a continuous and differentiable function; however, unlike previous methods, DGSM is explicitly designed to support gradient-based optimization, rather than statistical smoothing for visual quality. Given a shading point \(x\), we project it into the light coordinate system to obtain its depth \(z\) and the corresponding shadow map depth \(z_s\). Instead of a hard comparison between \(z\) and \(z_s\), DGSM evaluates visibility using a differentiable probability function, enabling effective optimization. This process is defined as:
\begin{align}
\label{eq:pdf}
{V = (z_s > z)} \rightarrow {V = P(z_s > z) = \text{Sigmoid}\big(k (z_s - z)\big)},
\end{align}
where the comparison $(z_s > z)$ is replaced by a probabilistic function $P(z_s > z)$, and $k$ controls the sharpness of the function.}

\revised{Finally, shadow-guided invisible geometry completion, our framework effectively reconstructs invisible regions, enabling physically plausible shadow synthesis.}

\subsection{\revised{Iterative material decomposition}}
\label{sec:material}
Urban scenes present significant challenges for decomposition due to sparse input views and complex lighting conditions. To address this, we introduce LMM to generate material supervision. LMM is the diffusion-based predictor that estimates per-pixel material properties (albedo, metallic, and roughness) from a single RGB image. These predictions provide useful guidance, but directly applying the model to real urban scenes can produce unrealistic results because the model is trained mainly on synthetic data. To reduce this synthetic-to-real domain gap, we propose an iterative material decomposition scheme. The key idea is to avoid treating the large material model predictions as fixed supervision. Instead, we repeatedly refine the predicted material maps by conditioning the model on the current Gaussian material estimates.


We first define the material model used in our framework. Each Gaussian primitive is associated with material properties $\mathbb{M}$, including albedo $\boldsymbol{a}$, roughness $\boldsymbol{r}$, and metallic $\boldsymbol{m}$. These properties are used in a simplified Cook–Torrance microfacet BRDF (Eqn.~\ref{eq:brdf}). \newrevised{Based on this material model, we present the general material supervision pipeline. For each training image $I_{\text{gt}}$, we use the LMM to predict the material maps $\mathcal{M}_0 = {A_0, M_0, R_0}$, formulated as $\mathcal{M}_0 = \mathrm{LMM}(I_{\text{gt}}, \sigma)$, where $\sigma$ denotes Gaussian noise. These maps are then used to supervise the optimization of the Gaussian material properties $\mathbb{M} = {a, m, r}$ through pixel-level material losses.}

\newrevised{Building on the general material supervision pipeline, we propose an iterative material decomposition scheme. Specifically, we perform an $N$-cycle update procedure with period $T$ ($N=3$, $T=6000$) to progressively refine the material maps. In the $n$-th cycle, the LMM-predicted material maps $\mathcal{M}_n$ supervise the optimization of $\mathbb{M}$ during the first $t-1$ iterations. At the $t$-th iteration, the updated Gaussian material properties are rasterized onto screen-space material maps $\mathcal{M}' = {A', M', R'}$. These maps are used as conditioning inputs to the LMM for diffusion-based inpainting, yielding refined material maps: $\mathcal{M}_{n+1} = \mathrm{LMM}(I_{\text{gt}}, \mathcal{M}')$, where $\mathcal{M}'$ replaces Gaussian noise as the condition. The refined material maps $\mathcal{M}_{n+1}$ are subsequently used as supervision in the next cycle.}

This alternating update process enforces consistency between the LMM outputs and the Gaussian representation, mitigates unrealistic material predictions arising from the synthetic-to-real domain gap, and enables robust material decomposition.

\revised{In implementation, we use WeatherDiffusion~\cite{zhu2025weatherdiffusion} as the LMM model to predict material supervision.}


\subsection{\revised{Physically-based lighting and shading}}
\label{sec:base}





Building upon the completion of invisible regions and scene material decomposition, we introduce physically-based lighting and shading to compute the final rendered image. Specifically, the final image is obtained by evaluating the rendering equation (Eqn.~\ref{eq:rendering}). 

\newrevised{Existing shading schemes include forward and deferred shading. Forward shading can produce blurred normals, as it computes shading per Gaussian and then alpha-blends the shaded colors; inconsistencies in normals across Gaussians lead to blurring. In contrast, deferred shading first alpha-blends the normals before shading, thereby avoiding such artifacts, as observed in GS-ROR$^2$~\cite{zhu2025gs-ror}. Therefore, we apply the deferred shading strategy, in which all required properties are first rasterized into G-buffers. The G-buffers include surface normals $N$, material properties $\mathcal{M} = \{A, R, M\}$, and visibility $V$, which are obtained by aggregating the corresponding properties of the Gaussian primitives.}

In order to capture the complex lighting in urban scenes, we decompose the outgoing radiance into a sunlight component $L_o^{\text{sun}}$, a skylight component $L_o^{\text{sky}}$. The sunlight is treated as directional light and modeled by two vectors: the sun intensity $S_i \in \mathbb{R}^3$ and the sun direction $S_d \in \mathbb{R}^3$. Based on Eq.~\ref{eq:rendering}, the $L_o^{\text{sun}}$ is calculated as:
\begin{align}
L_o^{\text{sun}} = f_r S_i V (S_d \cdot N),
\end{align}
where $f_r$ is the BRDF defined in Eq.~\ref{eq:brdf}.

For the skylight outgoing radiance \(L_o^{\text{sky}}\), we model skylight using an environment map and compute $L_o^{\text{sky}}$ through the split-sum approximation. Since urban scenes often exhibit complex geometry and self-occlusions, we introduce a learnable ambient occlusion attribute $\boldsymbol{ao}$ for each Gaussian primitive to approximate skylight occlusion. This allows the model to capture fine-scale geometric occlusions and improve detail fidelity.  Moreover, considering the indirect light within the scene, we apply a neural network \(\mathcal{U}\) to predict indirect outgoing radiance \(L_o^{\text{ind}}\). The network takes G-buffers—including \(N\), \(\mathcal{M}\), \(L_o^{\text{sun}}\), and \(L_o^{\text{sky}}\)—as inputs and outputs the indirect outgoing radiance \(L_o^{\text{ind}}\).  Finally, the outgoing radiance $L_o$ is computed as the sum of the each component: $L_o = L_o^{\text{sun}} + L_o^{\text{sky}} + L_o^{\text{ind}}$. We then apply gamma correction to $L_o$ and convert it to the RGB color space to generate the final shaded image $I$. For more details about the lighting and shading model, please refer to the supplementary materials Sec~\ref{sec:supp:light}.


\revised{In practice, to support relighting with adjustable illumination, we introduce two types of visibility modeling. Editable visibility $V_e$ is computed using shadow mapping (Sec.~\ref{sec:shadow}), enabling flexible relighting. In contrast, fixed visibility $V_f$ is derived from the Gaussian visibility properties $\boldsymbol{v}$ and is used for the NVS task.}

\subsection{Optimization strategy and loss functions}
\label{sec:training}
\revised{Our framework consists of optimizable parameters—Gaussian primitives, the environment map, the indirect light network $\mathcal{U}$, and the sunlight intensity $S_i$—as well as fixed parameters, including camera intrinsics/extrinsics (from COLMAP~\cite{schoenberger2016mvs, schoenberger2016sfm}) and sunlight direction $S_d$.} To train the framework, we use a two-stage strategy: the first stage focuses on Gaussian initialization, while the second stage enables relighting. The loss function $L_{\text{s1}}$ used in the first stage is defined as follows:
\begin{align}
\label{eq:loss_s1}
L_{\text{s1}} =& \lambda_\text{c} L_{\text{c}} + \lambda_{\text{ND}}L_{\text{ND}} + \lambda_{\text{N}} L_{\text{N}},
\end{align}
where $L_{\text{c}}$ denotes the RGB color loss from 3DGS~\cite{kerbl20233d}. $\lambda_{\text{ND}}$ represents the normal consistency loss between the Gaussian normal $N$ and the depth-derived normal $N_D$. $L_{\text{N}}$ is the normal regularization loss between the estimated normal $N'$ from a normal estimator~\cite{bae2024dsine} and $N$.

In the second stage, we introduce material properties and light parameters along with their corresponding loss functions. The loss functions $L_{\text{s2}}$ for the second stage is as follows:
\begin{align}
\label{eq:loss_s2}
L_{\text{s2}} &= L_{\text{s1}} + \lambda_\text{s} L_{\text{s}} + \lambda_{\text{Ve}} L_{\text{Ve}} + \lambda_{\text{Vf}} L_{\text{Vf}} + \lambda_{\text{M}} L_{\text{M}} + \lambda_{\text{G}}' L_{\text{G}}' + \lambda_\text{novel} L_{\text{novel}}.
\end{align}
Among them, $L_{\text{s}}$ is the shaded RGB color loss between the shaded image $I$ and the ground truth image $I_{\text{gt}}$. $L_{\text{M}}$ is the material loss enforcing material consistency between LMM-predicted materials (\( A', M', R' \)) and Gaussian materials. \newrevised{We use MTMT~\cite{chen2020MTMT} to estimate the ground-truth visibility $V'$ for supervision. $L_{\text{Vf}}$ denotes the binary cross-entropy (BCE) loss between the fixed visibility $V_f$ and $V'$. However, the ground-truth visibility is inconsistent across views, and directly supervising the geometry-related editable visibility would introduce geometric bias. Therefore, we use the cross-view consistent fixed visibility to supervise the editable visibility, defining $L_{\text{Ve}}$ as the BCE loss between $V_e$ and $V_f$.} $L_{\text{G}}'$ denotes a geometry regulation loss, used to refine the overall geometric structure of a scene from any aerial perspective. In addition, we introduce a novel-view loss $L_{\text{novel}}$ based on Difix3D to provide supervised training for invisible regions. The implementation of the losses is detailed in the supplementary materials.

\begin{figure*}
    \begin{center}
        \includegraphics[width=0.95\textwidth]{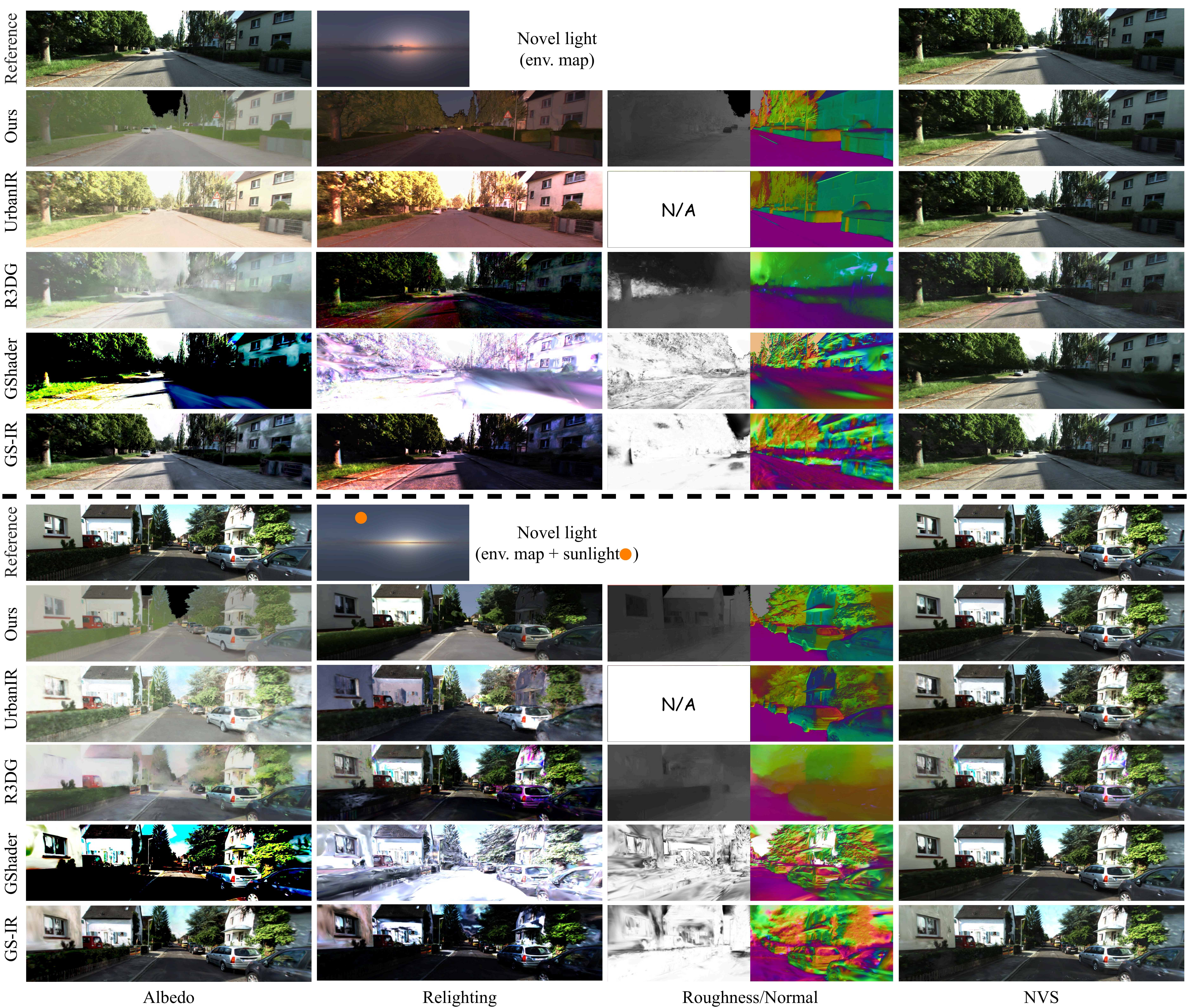}
    \end{center}
        \caption{\newrevised{Visual comparison of NVS, relighting, and material–lighting decomposition on the KITTI-360 dataset. Our method effectively mitigates the lighting bake-in issue during material decomposition, achieving more reliable relighting results while preserving high-quality novel view synthesis.}}
        \label{fig:eval:real_kitti360}
\end{figure*}

\begin{table*}[]
\caption{\revised{NVS quantitative evaluation on real-world datasets. Metrics highlighted in \textred{red} indicate the best performance, while those in \textgold{gold} indicate the second-best. Our method achieves the best results on most metrics, demonstrating high-fidelity scene reconstruction. With denser input views on the TandT100 dataset, our PSNR is slightly lower than that of GShader; however, when the input views are further sparsified in the TandT50 setting, our method attains the best performance. Moreover, due to the incorporation of the 3D completion model and DGSM, our method incurs higher computational overhead compared to Gaussian-based methods. Finally, we provide visual comparisons in Figs.~\ref{fig:eval:real_kitti360}, \ref{fig:eval:real_tandt}, and \ref{fig:eval:real_waymo}.}}
\label{tab:eval:real}
\centering
\resizebox{0.75\textwidth}{!}{%
\begin{tabular}{@{}c|ccc|ccc|ccc|ccc|lr@{}}
\toprule
 &
  \multicolumn{3}{c|}{KITTI-360} &
  \multicolumn{3}{c|}{Waymo} &
  \multicolumn{3}{c|}{TandT100} &
  \multicolumn{3}{c|}{TandT50} &
  \multicolumn{2}{c}{Time} \\ \cmidrule(l){2-15} 
\multirow{-2}{*}{Method} &
  PSNR &
  SSIM &
  LPIPS &
  PSNR &
  SSIM &
  LPIPS &
  PSNR &
  SSIM &
  LPIPS &
  PSNR &
  SSIM &
  LPIPS &
  \multicolumn{1}{c|}{Traing} &
  \multicolumn{1}{c}{Inference} \\ \midrule
GS-IR &
  \gold{22.73} &
  0.756 &
  0.219 &
  \red{32.44} &
  \gold{0.926} &
  \gold{0.150} &
  \red{20.72} &
  0.747 &
  0.201 &
  \gold{18.86} &
  0.671 &
  0.256 &
  \multicolumn{1}{l|}{\red{$\sim$1h}} &
  \red{$\sim$15 fps} \\
GShader &
  21.80 &
  \gold{0.780} &
  0.229 &
  29.89 &
  0.915 &
  0.192 &
  20.54 &
  \gold{0.776} &
  \gold{0.193} &
  18.70 &
  \gold{0.703} &
  \gold{0.239} &
  \multicolumn{1}{l|}{\gold{$\sim$2h}} &
  \gold{$\sim$13 fps} \\
R3DG &
  20.86 &
  0.734 &
  0.257 &
  27.79 &
  0.904 &
  0.180 &
  18.40 &
  0.664 &
  0.292 &
  17.50 &
  0.611 &
  0.342 &
  \multicolumn{1}{l|}{$\sim$2.5h} &
  $\sim$3 fps \\
UrbanIR &
  21.71 &
  0.778 &
  \gold{0.204} &
  28.37 &
  0.901 &
  0.199 &
  18.93 &
  0.694 &
  0.255 &
  16.49 &
  0.591 &
  0.387 &
  \multicolumn{1}{l|}{$\sim$10h} &
  $\sim$0.6 fps \\ \midrule
Ours &
  \red{23.24} &
  \red{0.824} &
  \red{0.160} &
  \gold{31.25} &
  \red{0.931} &
  \red{0.146} &
  \gold{20.63} &
  \red{0.792} &
  \red{0.178} &
  \red{19.19} &
  \red{0.741} &
  \red{0.206} &
  \multicolumn{1}{l|}{$\sim$4h} &
  $\sim$5 fps \\ \bottomrule
\end{tabular}%
}
\end{table*}

\section{Results}
\label{sec:results}
\subsection{Implementation details}
SRUG is implemented based on PyTorch~\cite{paszke2019pytorch}. In the first stage, we train a total of 30K steps. In the second stage, we train a total of 20K steps. All of our experiments are conducted on an RTX 3090 GPU.

\subsection{Experiment setups}
\paragraphNew{Dataset.}
We evaluate our method on four diverse datasets: two real-world driving scenes (KITTI-360~\cite{Liao2022KITTI-360} and Waymo \cite{sun2020scalability_waymo}), one real-world urban dataset (Tanks and Temple \cite{KnapitschTandT}), and one synthetic urban dataset. To assess performance under sparse-view conditions, we construct two subsets of the Tanks and Temple dataset by downsampling the number of training images: TandT100, with 100 training views, and TandT50, with 50 training views. \revised{For the synthetic dataset, we select 2 representative urban scenes and configure 3 different lighting conditions per scene—one used for training and the remaining 2 for validation. For more details on the datasets, please refer to Sec.~\ref{sec:supp:dataset} in the supplementary materials.}

\paragraphNew{Baseline.}
We select Gaussian-based relighting methods GS-IR~\cite{liang2024gsir}, Relightable 3D Gaussians (R3DG)~\cite{gao2025relightable}, GaussianShader (GShader)~\cite{jiang2024gsshader}, and a NeRF-based urban scene relighting method UrbanIR~\cite{lin2023urbanir} for comparison.

\subsection{Comparison}

We conduct comprehensive evaluations on both real-world and synthetic datasets. On real-world data, we assess NVS quality and present visual comparisons of material decomposition and relighting. On synthetic data, we quantitatively evaluate NVS, relighting, and material decomposition, complemented by visual comparisons. We use PSNR, SSIM~\cite{wang2004ssim}, and LPIPS~\cite{zhang2018lpips} to evaluate NVS, relighting, and albedo quality, and mean absolute error (MAE) for surface normals and roughness. Roughness results for UrbanIR are reported as N/A, as it does not model this property. We also report average training and inference times on real-world datasets to evaluate computational efficiency.

\paragraphNew{Real-world dataset.}
As shown in Table~\ref{tab:eval:real}, SRUG achieves superior performance across most NVS metrics. The slightly lower PSNR on Waymo can be attributed to our geometric regularization, which prioritizes geometric fidelity at the expense of appearance quality. Although the PSNR on TandT100 is comparatively modest, SRUG significantly outperforms competing methods on the more challenging TandT50 subset, demonstrating strong robustness under limited-view conditions.

\begin{figure*}
    \begin{center}
        \includegraphics[width=1.0\textwidth]{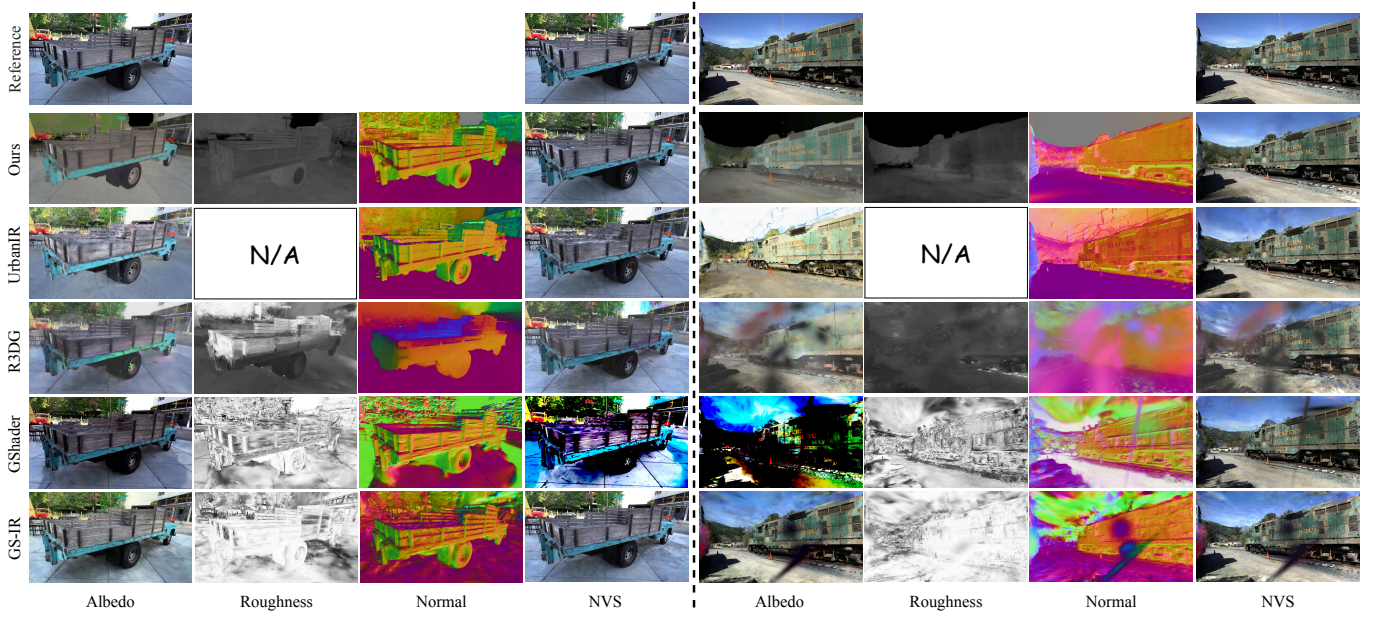}
    \end{center}
        \caption{Comparison of NVS and material decomposition on the TandT100 dataset (left part) and the TandT50 dataset (right part). }
        \label{fig:eval:real_tandt}
\end{figure*}

\begin{figure*}
    \begin{center}
        \includegraphics[width=0.95\textwidth]{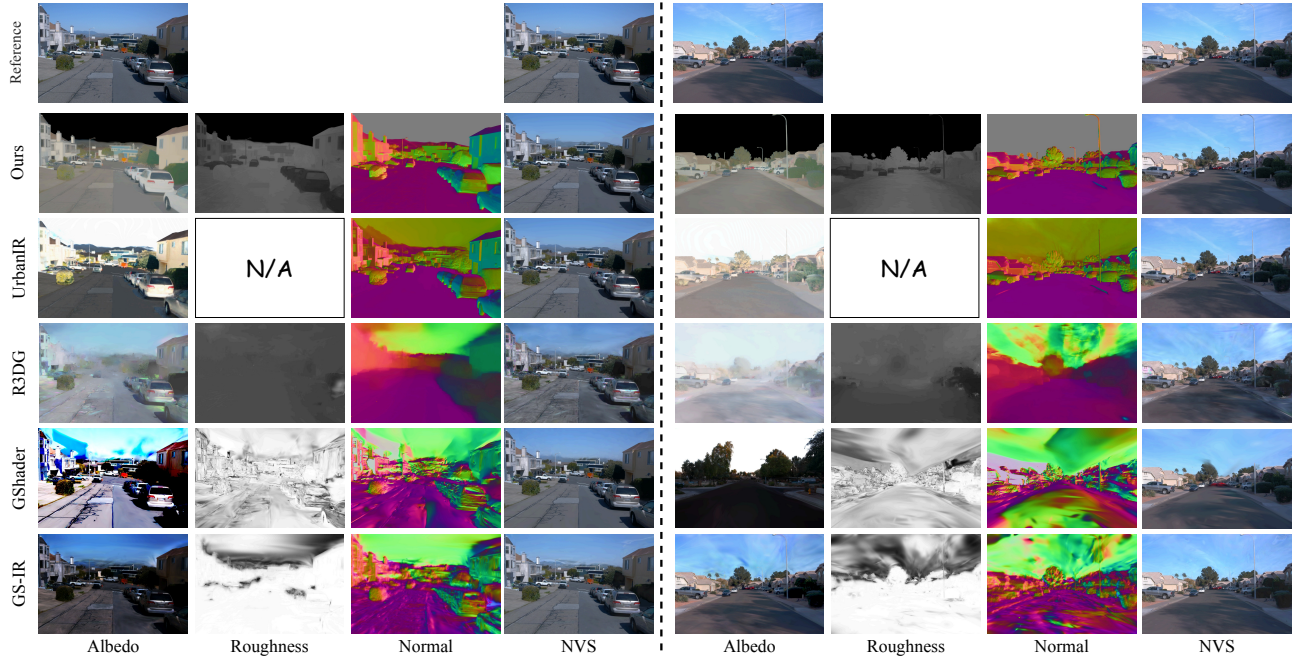}
    \end{center}
        \caption{Comparison of NVS and material decomposition on the Waymo dataset. Existing methods, including GS-IR, GShader, and R3DG, struggle to handle the complex illumination in urban scenes and suffer from severe lighting bake-in artifacts in the estimated albedo. While UrbanIR can effectively estimate materials in road regions, noticeable lighting bake-in persists in other parts of the scene, such as buildings. In contrast, our method achieves a more accurate and consistent separation of material and lighting across the entire scene.}
        \label{fig:eval:real_waymo}
\end{figure*}

Visual comparisons (Figs.~\ref{fig:eval:real_kitti360}, \ref{fig:eval:real_tandt}, and \ref{fig:eval:real_waymo}) further show that our material decomposition scheme enables more accurate separation of material properties, effectively mitigating lighting bake-in artifacts. Efficiency results reported in Table~\ref{tab:eval:real} indicate that, while the incorporation of the 3D completion model and the iterative material decomposition scheme increases training time relative to purely Gaussian-based methods, our approach still achieves a $2\times$ speedup over UrbanIR. Similarly, although DGSM introduces additional inference overhead, SRUG remains approximately $10\times$ faster than UrbanIR during inference.

\begin{figure*}
    \begin{center}
        \includegraphics[width=0.95\textwidth]{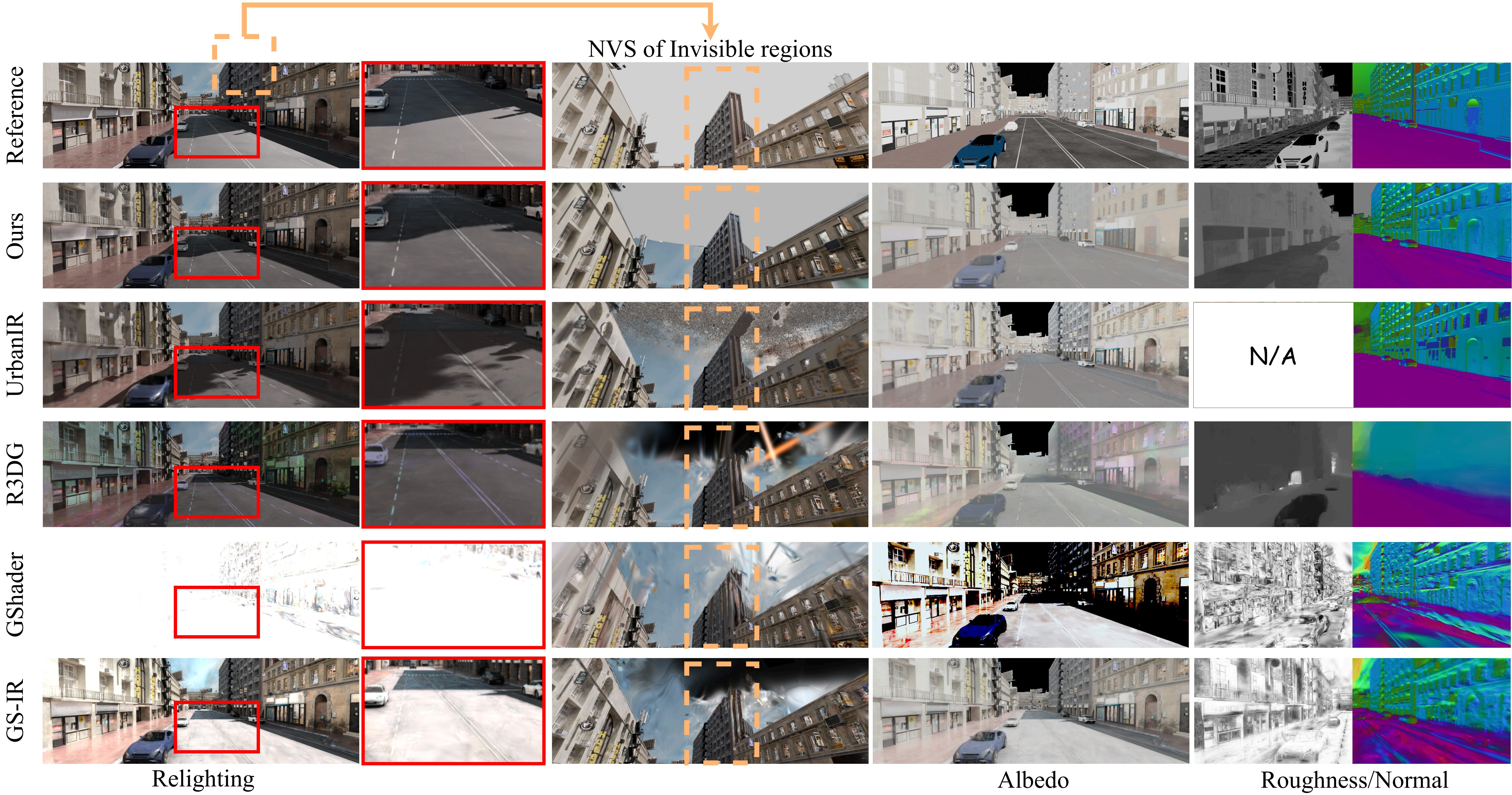}
    \end{center}
        \caption{\newrevised{Comparison of NVS, relighting, and material decomposition on the synthetic dataset. Under novel lighting conditions, our method achieves the most realistic relighting results. In contrast, UrbanIR introduces noticeable relighting artifacts, including inconsistent shadows and lighting bake-in effects, which degrade visual quality.}}
        \label{fig:eval:synthetic}
\end{figure*}

\paragraphNew{Synthetic dataset.}
Table~\ref{tab:eval:synthetic} summarizes the quantitative evaluation on the synthetic dataset, including NVS, relighting, and material decomposition metrics. SRUG achieves superior performance across all metrics, particularly in relighting and material decomposition. These quantitative improvements are further supported by the visual comparisons in Fig.~\ref{fig:eval:synthetic}, which demonstrate SRUG’s enhanced ability to decompose material properties and generate more accurate relighting compared to existing methods.

\begin{table}[]
\caption{\newrevised{NVS, material decomposition and relighting evaluation results on the synthetic dataset. Metrics in \textred{red} indicate the best, and in \textgold{gold} indicate the second-best. Our method achieves the best performance across most metrics, particularly excelling in material decomposition and relighting. For more intuitive results, we present the visual comparison in Fig.~\ref{fig:eval:synthetic}.}}
\vspace{-2mm}
\label{tab:eval:synthetic}
\centering
\resizebox{\linewidth}{!}{%
\begin{tabular}{@{}c|ccc|ccc|cll|c|c@{}}
\toprule
\multirow{2}{*}{Method} &
  \multicolumn{3}{c|}{Novel View Synthesis} &
  \multicolumn{3}{c|}{Relighting} &
  \multicolumn{3}{c|}{Albedo} &
  \multicolumn{1}{l|}{Roughness} &
  Normal \\ \cmidrule(l){2-12} 
        & PSNR & SSIM & LPIPS & PSNR & SSIM & LPIPS     & PSNR & SSIM & LPIPS & MAE & MAE \\ \midrule
GS-IR   & \gold{29.49} & \gold{0.905} & \gold{0.081} & 10.77 & 0.502 & 0.367 & 9.86 & 0.456 & \gold{0.397} & 0.181 & 0.305 \\
GShader & 28.44 & 0.902 & 0.098 & 3.77  & 0.327 & 0.754 & 11.50 & 0.451 & \gold{0.397} & 0.111 & 0.360 \\
R3DG   & 24.60 & 0.821 & 0.174 & 16.26 & 0.683 & \gold{0.273} & 10.96 & 0.457 & 0.513 & \gold{0.066} & 0.312 \\
UrbanIR & 27.19 & 0.856 & 0.161 & \gold{18.99} & \gold{0.716} & 0.286 & \gold{12.30} & \gold{0.477} & 0.418 & N/A   & \gold{0.204} \\ \midrule
Ours    & \red{30.10} & \red{0.927} & \red{0.079} & \red{20.56} & \red{0.810} & \red{0.160} & \red{14.78} & \red{0.515} & \red{0.392} & \red{0.035} & \red{0.186} \\ \bottomrule
\end{tabular}%
}
\end{table}

\begin{table}[h!]
\caption{\newrevised{The ablation study of the shadow-guided invisible geometry completion module on the synthetic dataset. Metrics in \textred{red} indicate the best, and in \textgold{gold} indicate the second-best. For more intuitive results, we also present the visual comparison in Fig.~\ref{fig:eval:ablation2}.}}
\vspace{-2mm}
\label{tab:eval:ablation2}
\centering
\resizebox{0.7\columnwidth}{!}{%
\begin{tabular}{@{}l|ccc@{}}
\toprule
Components & PSNR  & SSIM  & LPIPS \\ \midrule
baseline   & 18.17 & 0.782 & 0.176 \\
+ 3D completion model     & \gold{19.56} & \gold{0.799} & \gold{0.166} \\
+ DGSM shadow guidance & \red{20.56} & \red{0.810} & \red{0.160} \\ \bottomrule
\end{tabular}%
}
\vspace{-2mm}
\end{table}


\subsection{Ablation}

\begin{figure*}
    \begin{center}
        \includegraphics[width=0.80\textwidth]{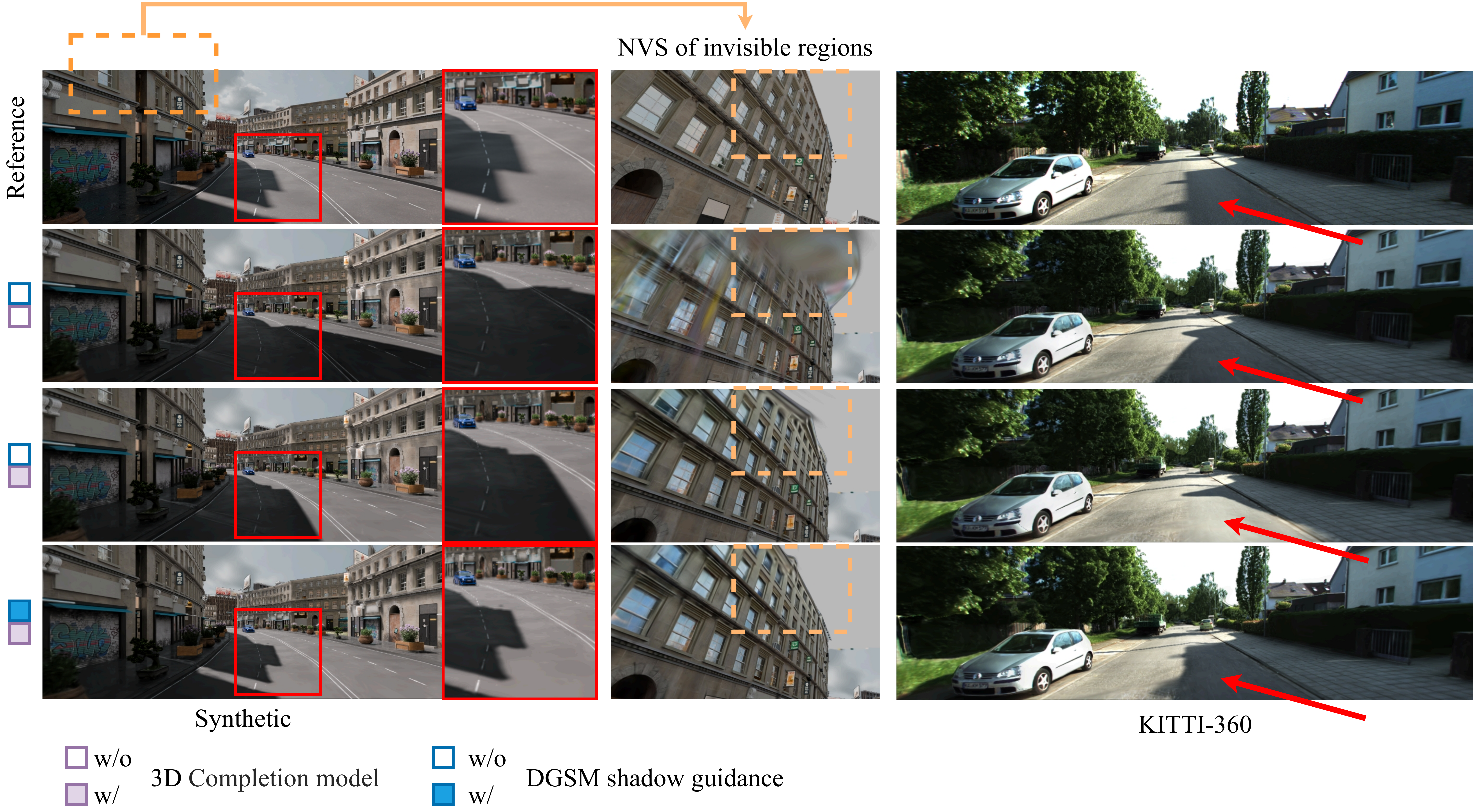}
    \end{center}
    \caption{\newrevised{We conduct an ablation study of the shadow-guided invisible geometry completion module on both synthetic datasets and KITTI-360. Incorporating the 3D completion model enables the recovery of geometry in invisible regions, while the proposed DGSM further improves geometric completion and shadow quality by leveraging shadow-based guidance, resulting in more physically plausible invisible geometry and shadow effects.}}
    \label{fig:eval:ablation2}
\end{figure*}

\revised{\paragraphNew{Shadow-guided invisible geometry completion.}
We evaluate the contribution of each component on the synthetic dataset through relighting tasks. We adopt standard non-differentiable shadow mapping as the baseline and progressively incorporate the 3D completion model and DGSM shadow guidance to form the complete model. The metrics in Table~\ref{tab:eval:ablation2} show that both the 3D completion model and DGSM improve relighting quality. As illustrated in Fig.~\ref{fig:eval:ablation2}, the baseline fails to reconstruct invisible regions, leading to noticeable shadow artifacts. Using only a 3D completion model still results in inconsistent shadows due to the lack of physical constraints. In contrast, incorporating DGSM shadow guidance suppresses these inconsistencies and achieves the best overall performance.}

\begin{figure}
    \begin{center}
        \includegraphics[width=0.95\columnwidth]{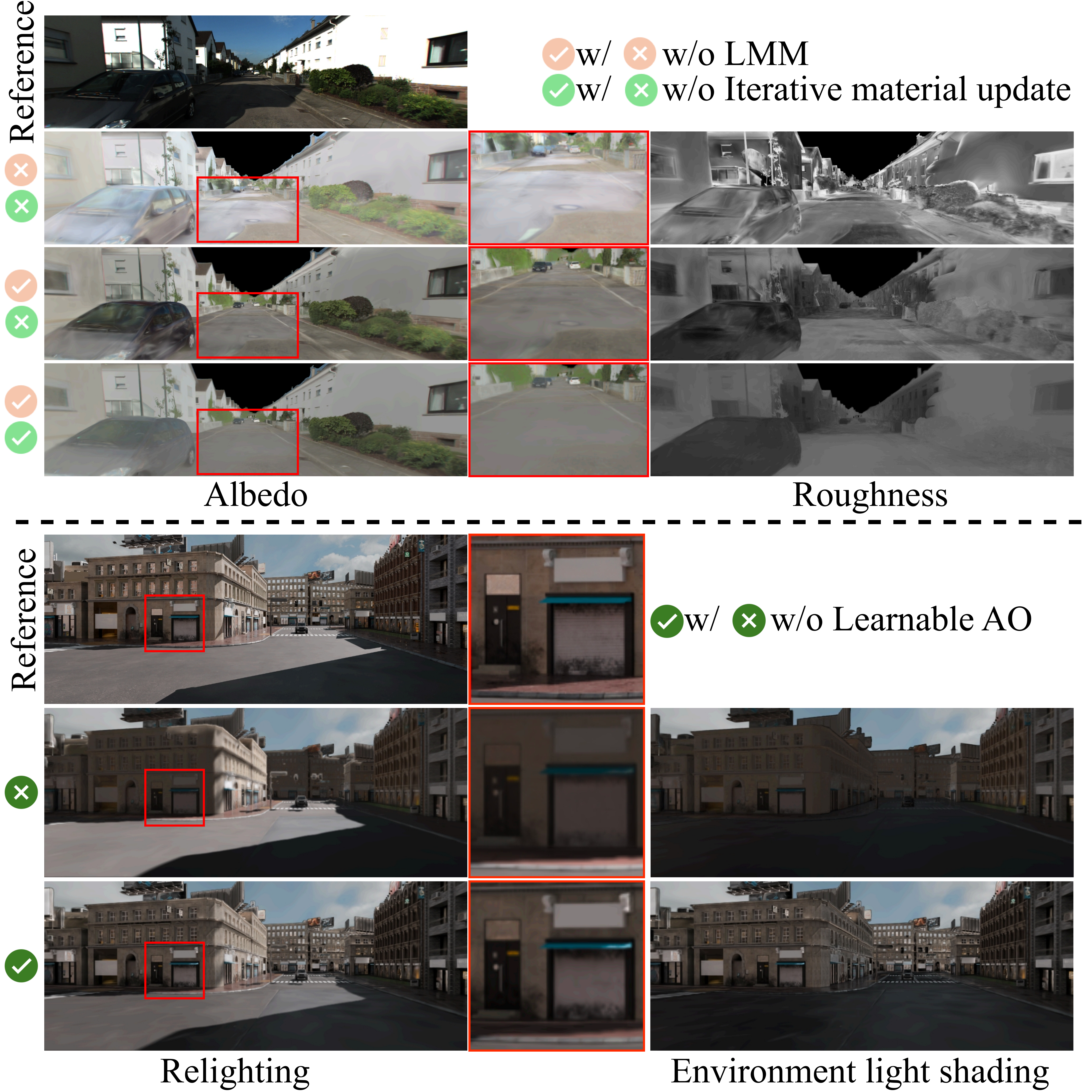}
    \end{center}
    \caption{\newrevised{Ablation studies of LMM-based material decomposition on the KITTI-360 dataset (upper part) and learnable AO relighting on the synthetic dataset (lower part). For LMM supervision, visual comparisons show that incorporating the LMM improves material decomposition and effectively reduces lighting bake-in artifacts, while the iterative material decomposition scheme produces more robust material estimates. For learnable AO, the visual results demonstrate that AO helps model subtle geometric occlusions, thereby enhancing rendering details, especially in shadow regions.}}
    \label{fig:eval:ablation1}
\end{figure}

\paragraphNew{Iterative material decomposition scheme.}
We progressively incorporate LMM supervision and iterative material decomposition to evaluate each component, starting from a baseline without LMM. Fig.~\ref{fig:eval:ablation1} illustrates their effectiveness. Without LMM, material properties and lighting are entangled, causing bake-in artifacts. LMM supervision mitigates this by material priors, but domain gaps between synthetic training data and real urban scenes leave artifacts under complex lighting. Our iterative material decomposition scheme effectively bridges this gap, yielding the most robust material estimates. Furthermore, even synthetic datasets can be affected by the domain gap due to variations in materials, lighting, and camera settings. Additional quantitative and visual ablations in Sec.~N.2 of the supplementary materials further demonstrate the effectiveness of the iterative material update scheme.

\revised{\paragraphNew{Learnable ambient occlusion.}
We validate the effectiveness of learnable ambient occlusion (learnable AO) through relighting experiments on the synthetic dataset. The quantitative results in Table~\ref{tab:eval:ablation3} show that AO effectively approximates skylight occlusion, leading to consistent improvements across multiple visual metrics. As shown in Fig.~\ref{fig:eval:ablation1}, learnable AO also helps model subtle geometric occlusions and enhances rendering details, especially in environment-light-dominated shadow regions.}

\begin{table}[ht!]
\caption{\newrevised{The relighting ablation study of the learnable AO on the synthetic dataset. Metrics in \textred{red} indicate the best.}}
\label{tab:eval:ablation3}
\centering
\resizebox{0.55\columnwidth}{!}{%
\begin{tabular}{@{}l|ccc@{}}
\toprule
Components & PSNR  & SSIM  & LPIPS \\ \midrule
w/o learnable AO   & 18.31 & 0.715 & 0.253 \\
Full model & \red{20.56} & \red{0.810} & \red{0.160} \\ \bottomrule
\end{tabular}%
}
\vspace{-3mm}
\end{table}



\section{Conclusion}
\revised{In this paper, we present SRUG, a novel framework for constructing relightable urban scenes from multi-view images or videos. SRUG leverages a shadow-guided 3D completion model to recover invisible regions, enabling physically plausible shadow synthesis. In addition, SRUG introduces an iterative material decomposition scheme that incorporates a large material model to provide material supervision and progressively achieve robust material decomposition. Building upon these components, SRUG enables reliable urban scene relighting through a physically-based lighting and shading model. Extensive quantitative and visual evaluations demonstrate that SRUG significantly advances the state-of-the-art in urban scene relighting compared to previous methods.}

\newrevised{\paragraph{Limitations and future work.} Although our iterative material decomposition scheme improves decomposition quality, it introduces multi-view inconsistencies in long-sequence inputs, resulting in blurred material estimates. Incorporating an LMM specifically designed for long sequences may alleviate this issue, which we leave for future work. Additionally, we plan to extend our method to larger-scale scenes, such as entire city blocks.}


\newpage


\bibliographystyle{ACM-Reference-Format}
\bibliography{paper}

\newpage

\clearpage
\appendix

\section{Background}
\label{sec:background}
\subsection{The rendering equation}
Classic surface rendering equation to formalize the outgoing radiance of the surface point $x$ with surface normal $\boldsymbol{n}$ as:
\begin{align}
\label{eq:rendering}
L_o\left(\omega_o, x\right) = \int_{\Omega} f_r\left(\omega_o, \omega_i, x\right) L_i\left(\omega_i, x\right) \left( \omega_i \cdot \boldsymbol{n} \right) \, d\omega_i,
\end{align}
where $f_r$ is the bidirectional reflectance distribution function (BRDF), $L_i$ and $L_o$ represent the incident and outgoing radiance in directions $\omega_i$ and $\omega_o$, and $\Omega$ represents the hemisphere above the surface. For BRDF, simplified Cook-Torrance microfacet BRDF~\cite{cook1982reflectance, walter2007microfacet} is a commonly used modeling method. It can be formulated as:
\begin{align}
\label{eq:brdf}
f_r = f_d + f_s = (1 - m)\frac{a}{\pi} + \frac{DFG}{4(\boldsymbol{n} \cdot \boldsymbol{l})(\boldsymbol{n} \cdot \boldsymbol{v})},
\end{align}
where the BRDF $f_r$ is divided into diffuse $f_d$ and specular $f_s$ components. Here, $\boldsymbol{m}$ denotes metallic, $\boldsymbol{a}$ is the albedo and $\boldsymbol{n}$ is the surface normal. And $\boldsymbol{v}$ and $\boldsymbol{l}$ are the view and light directions, respectively. The terms $D$, $F$, and $G$ correspond to the microfacet normal distribution function, Fresnel term, and shadowing-masking term.

\subsection{3D Gaussian splatting}
\label{sec:supp:3dgs}
3DGS represents 3D scenes as a set of 3D Gaussian primitives. Each primitive is modeled by a 3D mean $\boldsymbol{\mu}$ and a 3D covariance matrix $\bm{\Sigma}$, and is expressed as: $\boldsymbol{g}(\boldsymbol{x}) = e^{-\frac{1}{2}(\boldsymbol{x} - \boldsymbol{\mu})^\text{T} \bm{\Sigma}^{-1}(\boldsymbol{x} - \boldsymbol{\mu})}$, where the covariance matrix $\bm{\Sigma} = \bm{R} \bm{S} \bm{S}^\text{T} \bm{R}^\text{T}$ is derived from a rotation matrix $\bm{R}$ and a scaling matrix $\bm{S}$. To render the image, 3DGS first transforms the 3D Gaussian primitives into camera coordinates using a world-to-camera transformation matrix $\bm{W}$. Then, these primitives are projected onto the image plane through a local affine transformation $\bm{J}$, resulting in a 2D covariance matrix: $\bm{\Sigma}' = \bm{J} \bm{W} \bm{\Sigma} \bm{W}^{\text{T}} \bm{J}^{\text{T}}$. The final color $C$ is produced by alpha-blending the projected primitives. This process can be represented as:
\begin{align}
    C = \sum_{i=1}^{n} \boldsymbol{c}_i \boldsymbol{\alpha}_i \boldsymbol{g}^{2D}_{i}(\boldsymbol{x}_i) \prod_{j=1}^{i-1} (1 - \boldsymbol{\alpha}_j \boldsymbol{g}^{2D}_{j}(\boldsymbol{x}_j)),
\end{align}
where $i$, $j$ index the Gaussian primitives; $\boldsymbol{c}_i$ is the view-dependent color; $\boldsymbol{\alpha}_i$ is the opacity; $\boldsymbol{g}^{2D}$ is the Gaussian value derived from $\bm{\Sigma}'$; and $\boldsymbol{x}_i$ denotes the image-plane coordinate.

\subsection{RaDe-GS: Rasterizing Depth in Gaussian Splatting}
RaDe-GS~\cite{zhang2024rade} is built upon the 3DGS~\cite{kerbl20233d} representation. Its core idea is to explicitly model depth and surface normals in 3DGS, thereby improving geometric representation. By accurately defining the intersection between camera rays and 3D Gaussian ellipsoids, RaDe-GS obtains more precise depth and normal estimates. Compared with standard 3DGS, it provides more accurate geometric modeling, making it well-suited to our framework. Moreover, compared with 2DGS~\cite{huang20242dgs}, RaDe-GS retains the stronger visual expressiveness of 3D Gaussian primitives.

\section{More details about the datasets}
\label{sec:supp:dataset}
\paragraphNew{Synthetic dataset}
The rendering process for the synthetic dataset involves rendering high-resolution images along with corresponding material properties such as albedo, roughness, metallic, and normal maps. The dataset is designed to cover a wide range of urban environments and lighting conditions, ensuring sufficient diversity and complexity for evaluation. Specifically, we select 2 representative urban scenes as base 3D models and use Blender to place cameras and render multi-view images together with materials. During rendering, we employ the CYCLE engine to achieve realistic image synthesis. Lighting is modeled using a combination of environment maps and directional light (sunlight). For each scene, we render 3 different lighting configurations, using one for training and the remaining for evaluating relighting. In addition, for each scene, we record the direction of the directional light to initialize the sunlight direction parameter $S_i$.

\paragraphNew{Sunlight in the datasets}
In the real-world KITTI-360 dataset, the sunlight direction is computed from GPS metadata (following UrbanIR~\cite{lin2023urbanir}) and is kept fixed during training. For datasets in which the sunlight direction is not available, we do not use sunlight shadows as guidance in order to avoid reconstruction errors caused by inaccurate sunlight direction estimation. The sunlight intensity is treated as an optimizable parameter; it is initialized to 1, and is optimized during training.

Each scene covers approximately 100 meters of street environment and contains 120 rendered images, of which 12 are uniformly sampled at equal intervals for testing, while the remaining images are used for training.

\section{More details about the lighting and shading model}
\label{sec:supp:light}
For ambient occlusion, following standard practice, we multiply it by the diffuse reflection term induced by skylight to approximate the occlusion of skylight by the scene.

For indirect outgoing radiance, we use a neural network $\mathcal{U}$ to predict it, which may introduce a potential risk of overfitting to a specific scene and thus limit adaptability to novel lighting conditions. To address this concern, we design a specialized network architecture. Specifically, we adopt a convolutional network as the backbone, enabling the model to learn indirect lighting from local pixels and leverage the strong generalization capability of convolutional operations, which helps mitigate overfitting. In addition, we restrict the network depth to limit the scale of parameters, further reducing the risk of overfitting. Moreover, the relighting experiments demonstrate that the network adapts well to diverse lighting conditions.

Specifically, we adopt a 10-layer U-Net~\cite{ronneberger2015u} as the network backbone, consisting of four downsampling layers. Each downsampling layer uses a convolution with a kernel size of 3 and a stride of 2. The upsampling layers mirror the downsampling structure, except that bilinear interpolation is employed for upsampling. In addition, the network includes two convolutional layers at the input and output, which are used to ingest the input features and to predict the final indirect outgoing radiance, respectively.



\begin{table}[h!]
\caption{\newrevised{Relighting comparison between Gaussian-based ray tracing and DGSM.}}
\label{tab:eval:trac}
\centering
\resizebox{0.7\columnwidth}{!}{%
\begin{tabular}{@{}l|cccc@{}}
\toprule
Components & PSNR  & SSIM  & LPIPS & FPS \\ \midrule
DGSM    & {20.56} & \red{0.810} & {0.160} & \red{17} \\
3DGS-based ray-tracing & \red{20.60} & {0.808} & \red{0.158} & {8} \\ \bottomrule
\end{tabular}%
}
\end{table}

\section{Processing for sky regions}
During the construction of relightable scenes, floating Gaussians may appear in sky regions and degrade the relighting quality. To mitigate this issue, we explicitly mask the sky to suppress such floating Gaussians. Specifically, we first use a segmentation network~\cite{mmseg2020} to extract sky masks from the training images. We then apply these masks during training together with random background augmentation. Concretely, this step is formulated as a color loss between the image rendered with a random background and the ground-truth image composited with the same background.

We further treat the sky as an infinitely distant background and model its appearance with a sky texture. This texture takes an arbitrary direction vector as input and outputs the corresponding color. During training, we compute the viewing direction of each pixel from the camera parameters and sample the sky texture accordingly. The sampled sky color is then alpha-blended with the rendered image, and a color loss is computed against the original ground-truth image. The overall color loss is obtained by combining the losses from these two processes.

\newrevised{The above treatment of sky regions is based on UrbanIR~\cite{lin2023urbanir}. A limitation of this design is that the predicted sky masks may be inaccurate near sky-foreground boundaries, leading to some sky pixels being misclassified. These errors can introduce edge artifacts in the relighting results. If accurate sky masks are available in the dataset, such artifacts can be avoided.}

\section{More details on the LMM}
\label{sec:supp:LMM}
\newrevised{The LMM prediction process incurs a reasonable time overhead. On an RTX 3090 GPU, preprocessing each KITTI-360 scene ($\sim$120 images) takes roughly 5 minutes to obtain visibility and normal priors and 20 minutes to extract material priors. Furthermore, in iterative material decomposition, updating 120 images at a resolution of [1408, 376] requires approximately 10 minutes per cycle, and three cycles add $\sim$30 minutes.}

The diffusion inpainting process is a key component of our method. Given the material maps $\mathcal{M}$, obtained from Gaussian, as the conditioning input, we first apply the forward diffusion process by injecting noise for $t$ steps, resulting in noisy material maps $\mathcal{M}_t$. These noisy maps are then passed through the LMM, performing $t$ denoising steps to generate the refined material maps $\mathcal{M}'$. Diffusion inpainting leverages the material maps obtained from splatting as conditioning inputs to regenerate the material priors. Additionally, in our implementation, the inpainting process is performed iteratively, with an iteration period set to $T=6000$. And we found that setting the noise step to 600 (out of a total of 1000 diffusion steps) yields the best performance.

\section{More details on the BRDF model}
\label{sec:supp:brdf_details}
The BRDF characterizes how light reflects off a surface. A simplified version of the Cook-Torrance microfacet BRDF~\cite{cook1982reflectance, walter2007microfacet} is commonly used to model BRDF, as introduced in Sec.~3 of the main paper. This model separates BRDF $f_r$ into two components: a diffuse term and a specular term, expressed as:
$$
f_d = (1 - m)\frac{a}{\pi}, \quad f_s = \frac{DFG}{4(\boldsymbol{n} \cdot \boldsymbol{l})(\boldsymbol{n} \cdot \boldsymbol{v})}.
$$

\subsection{Diffuse component}
The diffuse reflectance depends solely on the surface albedo $\boldsymbol{a}$ and the metallic $\boldsymbol{m}$.

\subsection{Specular component}
The specular reflectance depends on three factors: the microfacet normal distribution function $D$, the Fresnel function $F$, and the geometric shadowing factor $G$.

\paragraphNew{The geometric shadowing factor $G$} accounts for the masking and shadowing caused by surface microgeometry. It is computed as:
\begin{align}
\label{eq:G}
G(\boldsymbol{n}, \boldsymbol{v}, \boldsymbol{l}, k) &= G_{\text{sub}}(\boldsymbol{n}, \boldsymbol{v}, k) \cdot G_{\text{sub}}(\boldsymbol{n}, \boldsymbol{l}, k), \\
G_{\text{sub}}(\boldsymbol{n}, \boldsymbol{v}, k) &= \frac{\boldsymbol{n} \cdot \boldsymbol{v}}{(\boldsymbol{n} \cdot \boldsymbol{v})(1 - k) + k},
\end{align}
where $\boldsymbol{n}$ is the surface normal, $\boldsymbol{v}$ and $\boldsymbol{l}$ are the view and light directions, and $k$ is a remapping parameter. For direct lighting, $k = \frac{(\alpha + 1)^2}{8}$; for image-based lighting, $k = \frac{\alpha^2}{2}$, with $\alpha = \boldsymbol{r}^2$, where $\boldsymbol{r}$ denotes the surface roughness.

\paragraphNew{The Fresnel function $F$} describes the ratio of reflected to refracted light at the interface. It can be approximated using the Fresnel-Schlick equation:
\begin{align}
\label{eq:F}
F(\boldsymbol{h}, \boldsymbol{v}, F_0) = F_0 + (1 - F_0)(1 - \boldsymbol{h} \cdot \boldsymbol{v})^5,
\end{align}
where $F_0$ is the reflectance at normal incidence (typically 0.04 for dielectrics), and $\boldsymbol{h} = \frac{\boldsymbol{v} + \boldsymbol{l}}{|\boldsymbol{v} + \boldsymbol{l}|}$ is the half-vector between the view and light directions.

\paragraphNew{The microfacet normal distribution function $D$} models the statistical distribution of microfacet orientations. One common choice is the GGX distribution, defined as:
\begin{align}
\label{eq:D}
D(\boldsymbol{n}, \boldsymbol{h}, \alpha) = \frac{\alpha^2}{\pi((\boldsymbol{n} \cdot \boldsymbol{h})^2(\alpha^2 - 1) + 1)^2}.
\end{align}

\section{More details about the framework training}
\label{sec:supp:training_details}
\subsection{Loss functions in the first stage}
The loss weights $[\lambda_\text{c}, \lambda_\text{ND}, \lambda_\text{N}]$ are set to $[1.0, 0.05, 0.05]$ in the first stage.
\paragraphNew{RGB color loss $L_\text{c}$} is the RGB color loss proposed in 3D Gaussian Splatting (3DGS)~\cite{kerbl20233d}, which consists of L1 loss and SSIM loss. We implement this loss according to the implementation method in 3DGS. $L_\text{c}$ can be expressed as:
$$
L_\text{c} = \lambda_1 \left\|I_r - I'\right\|_1 + \lambda_2 (1 - SSIM(I_r, I')),
$$
where $I_r$ is the image rendered using Gaussian color properties, $I'$ is the ground truth image, and $\lambda_1=0.8$ and $\lambda_2=0.2$ are hyperparameters controlling the relative weights of the L1 loss and SSIM loss, respectively. 

\paragraphNew{Normal consistency loss $L_\text{ND}$} is the normal consistency loss proposed by 2DGS~\cite{huang20242dgs}. This loss ensures that the Gaussian primitive locally approximates the actual object surface. $L_\text{ND}$ can be expressed as:
$$
L_{ND} = 1 - N \cdot N_{\text{D}},
$$
where $N$ represents the normal map from Gaussian, and $N_{\text{D}}$ is the depth-derived normal estimated by the gradient of the depth map. 

\paragraphNew{Normal regularization losses $L_\text{N}$} are designed to ensure that the depth and normal from Gaussian are consistent with the ground truth depth and normal. Specifically, we use the depth map and normal map estimated by the depth estimator~\cite{yang2024depth-anythingv2} and normal estimator~\cite{bae2024dsine} as supervision. $L_\text{N}$ can be expressed as:
$$
L_{N} =  1 - N \cdot N',
$$
where $N$ is the normal map from Gaussian, and $N'$ is the ground truth normal map.

Building upon $L_{\text{N}}$, we further introduce a bilateral smoothing loss to encourage normal smoothness and suppress abrupt artifacts. The bilateral smoothing loss $\mathcal{L}{_\text{b}}$ is defined as:
\begin{align}
\mathcal{L}_{\text{b}} = (M \odot \|\nabla N\|\text{exp}^{-\|\nabla I_{\text{gt}}\|}),
\end{align}
where $\nabla$ denotes the gradient operator and $I_{\text{gt}}$ is the ground-truth image. This loss is applied to the surface normals $N$, promoting material and geometric smoothness while preserving image-aligned discontinuities. The loss is further masked by $M$, which is predicted by a segmentation model~\cite{mmseg2020} and restricts smoothing to regions expected to be smooth, including roads and buildings.

\subsection{Loss functions in the second stage}
The loss weights $[\lambda_\text{s}, \lambda_{\text{M}}, \lambda_{\text{Vf}}, \lambda_{\text{Ve}}, \lambda_{G}']$ are 
\newline set to $[1.0, 1.0, 0.05, 10^{-2}, 0.02]$ in the second stage.
\paragraphNew{Shaded RGB color loss $L_{\text{shade}}$} supervises the shaded image $I$ using a combination of L1 and SSIM losses, similar to the RGB color loss $L_\text{c}$.

\paragraphNew{Material consistency loss $L_{\text{M}}$} is used to enforce consistency between the Gaussian-assigned material properties and those predicted by the LMM. Specifically, it ensures that the albedo, metallic, and roughness maps derived from the Gaussians match the corresponding outputs from LMM. The loss is computed as:
$$
L_{\text{M}} = \left\| A - A' \right\|_1 + \left\| M - M' \right\|_1 + \left\| R - R' \right\|_1,
$$
where $A'$, $M'$, and $R'$ are the albedo, metallic, and roughness maps predicted by LMM, and $A$, $M$, and $R$ are the corresponding material maps obtained from Gaussian.

\paragraphNew{Visibility loss} $L_{\text{Vf}}$ is the loss between the fixed visibility $V_f$ from Gaussian and the ground truth visibility $V'$ estimated by the visibility estimator~\cite{chen2020MTMT}, and $L_{\text{Ve}}$ is between the editable visibility $V_e$ and $V_f$. These are computed as:
$$
L_{\text{Vf}} = \text{BCE}(V_f, V'), \quad L_{\text{Ve}} = \text{BCE}(V_e, V_f).
$$

\paragraphNew{Geometry regulation loss $L_{\text{G}}'$} is the depth distortion loss proposed by 2DGS~\cite{huang20242dgs} to encourage Gaussian primitives and remain compact and mitigate the adverse effects of floating Gaussian primitives on shadow effects. Specifically, we extend the depth distortion loss $L_{\text{dist}}$ to arbitrary aerial perspectives. We randomly sample camera perspectives to render the scene and compute the depth distortion loss. This makes the Gaussian distribution more compact and avoids artifacts.

\paragraphNew{The novel-view loss $L_{\text{novel}}$} is special. In each training iteration, we select an image from the  Difix3D-completed view, and we compute the losses $L_c$. The loss function is added to the total loss with a weight of 0.2.

In addition, the selection of novel views is completely random. Specifically, the random transformations include translations along the camera plane of ±0.1 units forward/backward and ±0.1 units left/right, rotation around the camera’s central axis of ±90°, and rotation along the camera’s pitch angle within [0°, 30°]. The camera intrinsic parameters remain unchanged during these transformations. During training, the new viewpoint dataset is updated every 2,000 steps. For all training cameras, these transformations generate new cameras, from which images are rendered and then completed using Difix3D. The completed images, together with the corresponding new cameras, form the novel-view dataset. Additionally, we apply a segmentation model~\cite{mmseg2020} to segment the sky region in the completed images, ensuring proper handling of the sky in the novel view data.

\subsection{Gaussian densification and pruning}
\paragraphNew{In the first stage,} we adopt the Gaussian densification and pruning strategy following Rade-GS~\cite{zhang2024rade}.
\paragraphNew{In the second stage,} densification is entirely guided by gradient backpropagation from the differentiable shadow mapping, with a gradient threshold set to $2 \times 10^{-5}$ and update interval set to 500 iterations, while other settings remain unchanged. The densification and pruning operations continue until 10000 iterations.



\section{Discussions about GS-ID and InvRGB+L.}
\label{sec:supp:gs_id}
GS-ID~\cite{du2024gsid} is related to our method in that it leverages a large material model (LMM) as a prior to improve material decomposition. However, GS-ID directly applies LMM without accounting for the domain gap between synthetic and real-world data. In practice, this is equivalent to removing the iterative update mechanism in our framework. In contrast, we adopt an iterative update scheme that progressively refines material maps, enabling adaptive correction and improved decomposition accuracy. This iterative process ensures that the material prior remains consistent and reliable in real-world scenarios, leading to more robust material decomposition. Moreover, GS-ID employs shadow mapping only at inference time to produce visually plausible relighting, whereas our method incorporates shadow mapping directly into the optimization process. By using shadow-based supervision during training, our framework learns to generate more realistic shadows and achieves higher-quality relighting results.

InvRGB+L~\cite{chen2025invrgb+} improves albedo decomposition by introducing LiDAR as an additional prior, which constitutes the key difference from our approach. In contrast, our method relies solely on the LMM as a prior, without requiring any extra sensor modalities.

\section{Discussion about Instruct-NeRF2NeRF.}
\label{sec:supp:gs_id}
Instruct-NeRF2NeRF~\cite{haque2023instruct} introduces an iterative strategy for high-quality NeRF scene editing. Inspired by this idea, we propose an iterative update scheme in material space to mitigate the domain gap between synthetic and real-world data in LMM predictions. Unlike Instruct-NeRF2NeRF, our objective is material decomposition from RGB images rather than scene editing. Due to the fundamental difference between RGB images and material representations, we do not update individual images; instead, we update the entire training set at each iteration. Moreover, to achieve accurate material estimation, the diffusion noise level $t$ is progressively reduced across iterations, rather than randomly sampled from a fixed range as in Instruct-NeRF2NeRF.

\section{Discussion on DGSM and Gaussian-based ray tracing}
Gaussian-based ray tracing can compute the opacity at a given point along a specified direction. In visibility estimation, this opacity can be directly interpreted as visibility, and the process is fully differentiable. However, this approach suffers from low computational efficiency. First, it requires computing intersections between each Gaussian and the ray, which is costly, especially in large-scale outdoor scenes. Second, ray tracing must be performed for all pixels in every frame, further increasing computational overhead, particularly for continuous sequences.

\newrevised{In contrast, our DGSM follows the shadow mapping paradigm. For scenes with fixed lighting, only a single rasterization is required for initialization, and subsequent frames involve only pixel projection and sampling, yielding a substantial performance advantage over ray tracing. We replace DSM with 3DGS-based ray tracing and conduct relighting experiments on the synthetic dataset. The results in Table~\ref{tab:eval:trac} show that DSM achieves higher efficiency with comparable rendering quality.}

\subsection{Discussion on two-stage training strategy}
Two-stage training is necessary in our framework. As a foundation for relighting, we first reconstruct the scene from the training data. Building on this initial reconstruction, we then recover invisible regions, decompose scene material properties, and construct a relightable representation. Consequently, the initialization provided by the first stage is essential; removing it leads to unstable reconstruction and prevents convergence to meaningful results. Moreover, two-stage training is a widely adopted strategy and has been used in many prior works~\cite{liang2024gsir, gao2025relightable}.

\section{Differences between our method and improved shadow mapping techniques}
\label{sec:supp:DGSM}
Our differentiable shadow mapping (DGSM) draws inspiration from advanced shadow mapping techniques~\cite{donnelly2006variance_vsm, gumbau2011shadow_gsm}. While these approaches also incorporate probabilistic modeling into traditional shadow mapping, their primary goal is to generate soft shadows by probabilistically accounting for neighboring pixels within the shadow map. In contrast, our method leverages probability not for soft shadow synthesis, but to enable differentiable shadow mapping. This formulation allows gradients to be propagated through the shadow computation, thereby supporting optimization of the 3D scene representation via backpropagation.

\section{More experiments}

\subsection{More ablation experiments about the iterative material decomposition scheme}
We also provide quantitative and visual ablation studies on the iterative material update scheme using synthetic data. Although all experiments are performed on synthetic datasets, domain gaps—such as variations in material properties, texture complexity, or camera configurations—can still pose challenges for material decomposition. Our iterative material update scheme effectively mitigates these issues, improving both consistency and accuracy in material decomposition. In addition, to evaluate the impact of the number of iterations on material decomposition, we conduct ablation experiments with $N = 1, 2, 3$ to assess how the iteration count affects decomposition quality.

\begin{table}[h!]
\caption{\newrevised{The quantitative ablation experiment of the iterative material update scheme on the synthetic dataset.}}
\label{tab:LMDM_syn}
\centering
\resizebox{0.8\columnwidth}{!}{%
\begin{tabular}{l|ccc|c}
\hline
\multirow{2}{*}{Method}       & \multicolumn{3}{c|}{Albedo} & Roughness \\ \cline{2-5} 
                              & PSNR    & SSIM    & LPIPS   & MAE       \\ \hline
w/o LMM                       & 14.35   & 0.512   & 0.400   & 0.062     \\
w/o iterative material update & 13.95   & 0.508   & \red{0.369}   & \gold{0.036}    \\
N = 1                         & 14.58   & \gold{0.514}   & \gold{0.378}  & \red{0.035}     \\
N = 2                         & \gold{14.74}   & \red{0.515}   & 0.384   & \red{0.035}     \\
N = 3 (Full model)            & \red{14.78}   & \red{0.515}   & 0.392   & \red{0.035}     \\ \hline
\end{tabular}%
}
\end{table}

\begin{figure}[h!]
    \begin{center}
        \includegraphics[width=0.95\linewidth]{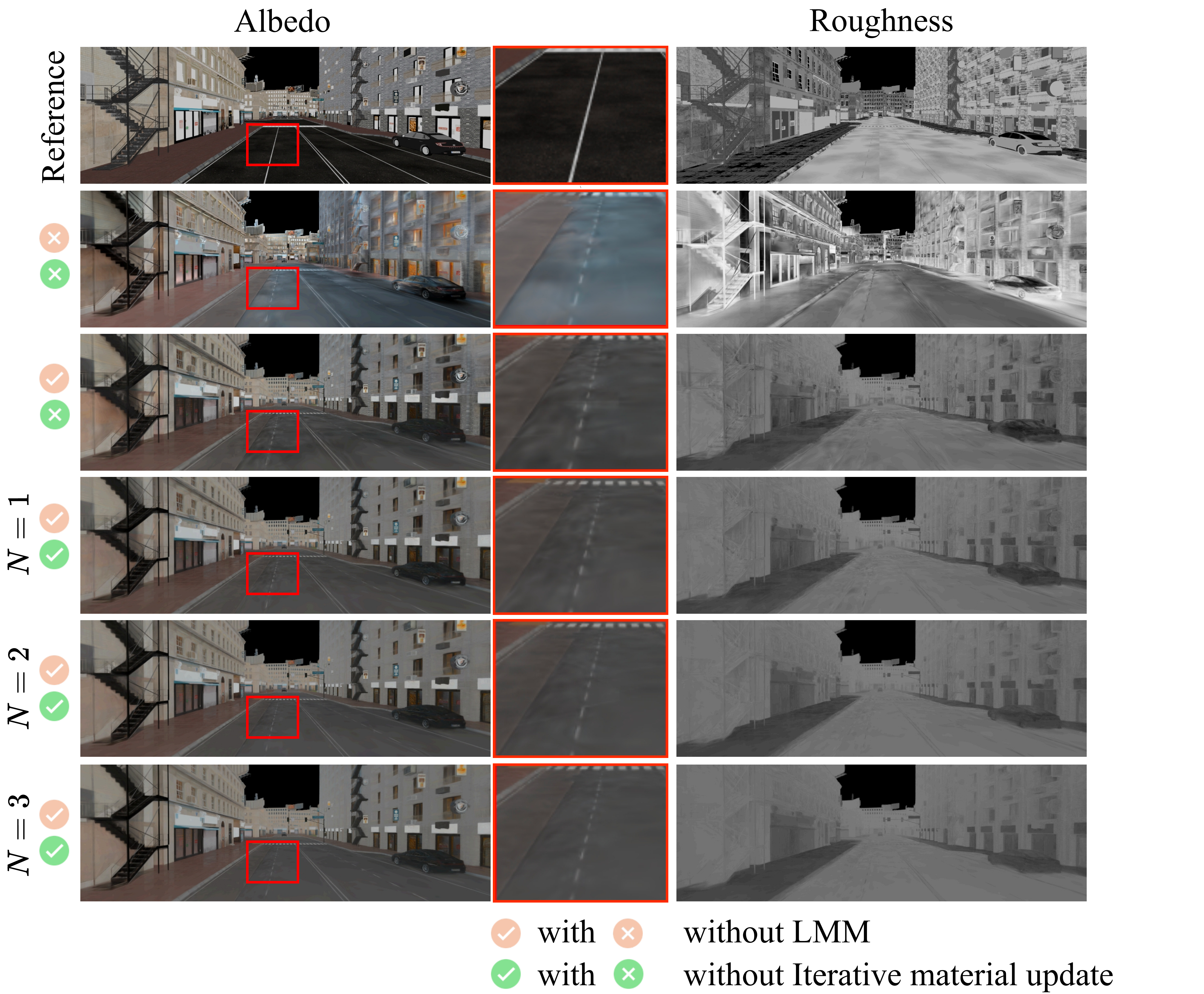}
    \end{center}
        \caption{\newrevised{Visual ablation of the iterative material update scheme on the synthetic dataset. Omitting the LMM or iterative updates introduces artifacts, whereas our iterative update scheme produces more robust materials. In our implementation, we set the iteration count $N=3$, which reduces artifacts while avoiding overly smooth material estimates.}}
        \label{fig:vis_LMDM}
\end{figure}

The quantitative results in Table~\ref{tab:LMDM_syn} indicate that omitting the LMM prior or using LMM without the iterative update scheme leads to artifacts in the estimated materials and degrading decomposition quality, especially for albedo. Our iterative update scheme effectively mitigates the artifacts, yielding more robust material decomposition. The visual results in Fig.~\ref{fig:vis_LMDM} further support this observation.

Additionally, as discussed in Sec.~5, due to multi-view inconsistencies of the LMM over long sequences, increasing the iteration count of update iterations tends to produce smoother materials, reflected in higher LPIPS metrics. Conversely, using fewer iterations results in more noticeable visual artifacts. Balancing smoothness and artifact suppression, we set the number of iterations to $N=3$, ensuring materials are neither overly smooth nor visually degraded.

\subsection{Dynamic scenes}
We add a demonstration of dynamic scenes in the supplementary video. Specifically, we insert a vehicle with material properties into the scene and manipulate its motion. However, it should be noted that our framework does not incorporate temporal modeling and therefore cannot directly handle dynamic input scenarios. Extending our method to dynamic scenes could be achieved by replacing the Gaussian initialization with existing dynamic Gaussian representations, which we leave for future work.

\end{document}